\documentclass[runningheads]{llncs}
\usepackage[T1]{fontenc}
\usepackage{graphicx}
\usepackage{xcolor,enumitem}
\usepackage{cite}

\usepackage{orcidlink}
\setenumerate[1]{itemsep=0pt,partopsep=0pt,parsep=0pt,topsep=0pt,leftmargin=0pt}
\setitemize[1]{itemsep=0pt,partopsep=0pt,parsep=0pt,topsep=0pt,leftmargin=12pt}
\setdescription{itemsep=0pt,partopsep=0pt,parsep=0pt,topsep=0pt,leftmargin=0pt}
\usepackage{enumitem,booktabs,multirow,array,pifont}
\newcommand{\etal}{\textit{et al.}}
\newcommand{\eg}{\textit{e.g.}}
\begin{document}
\title{ColorGPT: Leveraging Large Language Models \\ for Multimodal Color Recommendation}
\titlerunning{ColorGPT: LLMs for Color Recommendation}
%
\author{
Ding Xia\inst{1,2}\orcidlink{0000-0002-4800-1112} \and
Naoto Inoue\inst{2}\orcidlink{0000-0002-1969-2006} \and
Qianru Qiu\inst{2}\orcidlink{0009-0007-4562-6014} \and
Kotaro Kikuchi\inst{2}\orcidlink{0000-0003-1747-5945}
}
%
\authorrunning{D. Xia et al.}

%
\institute{The University of Tokyo, Tokyo, Japan \\
\email{dingxia1995@gmail.com}\\ \and
CyberAgent AI Lab, Tokyo, Japan\\
\email{\{inoue\_naoto,qiu\_qianru,kikuchi\_kotaro\_xa\}@cyberagent.co.jp}}

\maketitle              
%

\begin{abstract}
\looseness=-1
    Colors play a crucial role in the design of vector graphic documents by enhancing visual appeal, facilitating communication, improving usability, and ensuring accessibility. In this context, color recommendation involves suggesting appropriate colors to complete or refine a design when one or more colors are missing or require alteration. Traditional methods often struggled with these challenges due to the complex nature of color design and the limited data availability.
    In this study, we explored the use of pretrained Large Language Models (LLMs) and their commonsense reasoning capabilities for color recommendation, raising the question: \textbf{Can pretrained LLMs serve as superior designers for color recommendation tasks?} To investigate this, we developed a robust, rigorously validated pipeline, \texttt{ColorGPT}, that was built by systematically testing multiple color representations and applying effective prompt engineering techniques. Our approach primarily targeted color palette completion by recommending colors based on a set of given colors and accompanying context. Moreover, our method can be extended to full palette generation, producing an entire color palette corresponding to a provided textual description.
    Experimental results demonstrated that our LLM-based pipeline outperformed existing methods in terms of color suggestion accuracy and the distribution of colors in the color palette completion task. For the full palette generation task, our approach also yielded improvements in color diversity and similarity compared to current techniques.

\end{abstract}

\section{Introduction}
\label{sec:intro}

Color is a fundamental element in visual media, playing a pivotal role in graphic design by evoking specific emotions in viewers and significantly influencing their comprehension of the media~\cite{kaya2004relationship,8978038,8978102}. For the color recommendation task, once the overall structure of a design is fixed, designers typically need to adjust colors to ensure harmony and alignment with the semantic concepts of the design. This complex process requires a deep understanding of color theories and heuristics, making it difficult for novice designers.

Various approaches to color modeling have been investigated to support such designers. These approaches can be classified into heuristic-based~\cite{matsuda1995color} and machine-learning-based methods~\cite{odonovan2011,kita2016aesthetic,qiu2023color,bahng2018coloring,maheshwari2021generating,qiu2022intelligent,qiu2023multimodal}.
While heuristic methods often failed to capture complex priors, machine-learning methods typically relied on labor-intensive data collection processes, limiting their scalability.
One of the most remarkable and promising trends in tackling data inefficiency was to utilize priors embedded in the pretrained LLMs, as investigated in knowledge graph~\cite{bosselut2019comet} and layout domains~\cite{feng2024layoutgpt}.
We can also see similar trends in color research, with some studies obtaining the semantic description of a single color without expensive manual annotations~\cite{abdou2021can,kawakita2024gromov,loyola2023perceptual,mukherjee2024large}.

Despite significant advances in LLMs, their application in color recommendation tasks remained underexplored. This raises a critical research question: \textbf{Can pretrained LLMs serve as superior designers for color recommendation tasks?} To answer this question, we need two main steps: First, we investigated effective color representations and prompt structures. Based on these insights, we developed a robust, rigorously validated pipeline tailored for the task. Second, we assessed the superiority and effectiveness of our approach across various types of LLMs.

We introduced \texttt{ColorGPT}, a robust and thoroughly validated pipeline that leverages the commonsense reasoning capabilities of LLMs to tackle the complexities of color design in graphic documents. Our approach was evaluated on two primary tasks: color palette completion and full palette generation. Full palette generation entails creating a harmonious set of colors based solely on textual descriptions, while color palette completion involves suggesting appropriate colors for a vector graphics document by considering both existing colors and the accompanying textual context.
We compared our approach with existing methods, and our results demonstrated that \texttt{ColorGPT} can effectively harness LLMs' commonsense reasoning to address complex, real-world color design challenges. Additionally, our analysis evaluated the impact of different color representations and prompts. Our findings indicated that the choice of color representation significantly influences performance and that prompts with similar in-context exemplars yield better results. 

Our contributions are summarized as follows:
\begin{itemize}
    \item We presented \texttt{ColorGPT}, a robust and thoroughly validated pipeline that leveraged the commonsense knowledge of large language models for color design tasks involving multi-modal data.
    \item We conducted extensive experiments on two color recommendation tasks, including color palette completion and full palette generation, confirming the superior performance and effectiveness of \texttt{ColorGPT}.
    \item We evaluated various component variants, including different color representations and prompt structures, and analyzed their impact on the outcomes.
\end{itemize}

\section{Related Works}
\label{sec:relate}

\subsection{Color Palette Completion}
Color palette completion aims to suggest appropriate colors that harmonize with existing palettes and align with human aesthetic preferences. Traditional methods~\cite{shimizu2003interactive,odonovan2011,kita2016aesthetic} relied on hand-crafted features and regression models to predict new colors based on given palettes.
Later, deep learning methods have become prominent in this field. Kim~\etal~\cite{kim2022colorbo} developed a model that predicts and recommends colors based on seed colors, leveraging a word-embedding-like approach. 
Yuan~\etal~\cite{yuan2021infocolorizer} utilized Variational AutoEncoder with Arbitrary Conditioning (VAEAC) to dynamically recommend color palettes based on user-defined conditions. 
Kikuchi~\etal~\cite{kikuchi2023generative} proposed a new dataset consisting of mobile e-commerce web pages and introduced several Transformer-based methods for generative colorization.
Qiu~\etal~\cite{qiu2022intelligent,qiu2023color} proposed Transformer-based models for color recommendation in landing pages and vector graphic documents, and they 
further improved the performance of the color recommendation method by incorporating the information of text contents and image labels~\cite{qiu2023multimodal}.
Their work utilizes multiple color palettes extracted from multiple design elements in the Crello dataset~\cite{yamaguchi2021canvasvae} and recommends colors corresponding to the existing ones in multiple palettes.
Providing multiple palettes for a graphic document with various elements presents a challenging task that further explores the capabilities of LLMs in color recommendation and understanding.

\subsection{Full Palette Generation}
\looseness=-1
Full palette generation is a special case of text-aware color recommendation, involving the creation of an entire color palette aligned with a given text.
Some developers offer online encyclopedias that provide a comprehensive resource to identify and understand various color names~\cite{colorname2024,colorlover2024,adobecolor2024}. These resources are valuable for designers and researchers working with color theory.
Early studies on full palette generation often relied on predefined palettes based on specific themes, such as magazine cover design~\cite{jahanian2013recommendation,yang2016automatic}. These methods were limited by their lack of flexibility and inability to adapt to varying contexts.
Later, neural network models were introduced to predict colors for specific design elements~\cite{kawakami2016character,monroe2017colors,kikuchi2023generative}.
Maheshwari~\etal~\cite{maheshwari2021generating} and Bahng~\etal~\cite{bahng2018coloring} proposed conditional generative models for generating color palettes for image colorization using text inputs like `warm sunshine' and `cute dog'. Their work introduced the Palette-And-Text (PAT) dataset, which offers a benchmark in this field. 
Recently, Qiu~\etal~\cite{qiu2023multimodal} proposed a multimodal masked color model using CLIP-based text representations to connect semantic content with palette suggestions, improving the performance of color generation tasks.
The generation of color palettes can generally reflect LLMs' ability to understand the combination of multiple colors. This perfectly serves as one of our tasks to evaluate LLMs' ability to recommend colors.

\subsection{LLMs in color-related tasks}
Recent advancements in LLMs have shown great potential in design tasks~\cite{feng2024layoutgpt,jia2023cole}.
Abdou~\etal~\cite{abdou2021can}
and Loyola~\etal~\cite{loyola2023perceptual} studied whether the topological structure of color name embedding in LLMs is grounded in the world, such as RGB space, CIELAB space and etc.
Kawakita~\etal~\cite{kawakita2024gromov} and Mukherjee~\etal~\cite{mukherjee2024large} answered the question of whether LLMs have color-concept associations similar to humans. These works imply a strong fine-item-level structural correspondence between color-neurotypical human participants and the recent GPT models.
However, all the aforementioned approaches focus on analyzing individual colors. The exploration of color palettes, which is apparently more challenging, remains limited.

Some color-related tasks have been applied to LLMs recently~\cite{shi2023nl2color,hou2024c2ideas}. Shi~\etal~\cite{shi2023nl2color} refine color palettes for charts based on textual inputs. Hou~\etal~\cite{hou2024c2ideas} recommend some color properties, including palettes, given the user intention.
Although previous works have demonstrated the application of LLMs for palette recommendation, their evaluations typically rely on subjective studies. In contrast, our study is focused on quantitative assessment, elucidating how subtle choices in implementation affect performance and to what extent they can match conventional data-driven approaches.

\section{ColorGPT}
\label{sec:method}
To ensure the robustness and effectiveness of our \texttt{ColorGPT} pipeline (Fig.\ref{fig:pipeline_visualization}), we carefully designed the system by focusing on three key components: color representations, document description structure, and prompt construction.

\subsection{Color Representations}
\label{sec:method_color}
The choice of color representation can significantly influence the method's performance. Different formats, such as words, hexcodes, RGB values, and CIELAB, each have distinct characteristics that could affect the results. However, there are no standard color representations. Common formats used in previous works are as follows:

\begin{itemize}
    \item \textbf{Word}~\cite{loyola2023perceptual}: Descriptive color words such as ``white'' or ``blue,'' which are intuitive to humans.
    \item \textbf{Hexcode}~\cite{kawakita2024gromov}: Colors are represented in the form ``\#RRGGBB'', where `RR', `GG', and `BB' are hexadecimal values for the red, green, and blue components, respectively. For example, white is ``\#ffffff''.
    \item \textbf{RGB}~\cite{kawakita2024gromov}: A triplet \([R, G, B]\), where \(R, G, B \in [0, 255]\), representing the intensity of the red, green, and blue channels. White is represented as \([255, 255, 255]\).
    \item \textbf{CIELAB}~\cite{abdou2021can,qiu2023color,qiu2023multimodal}: A perceptually uniform color space, where colors are described by \((L^*, a^*, b^*)\), with \(L^*\) representing lightness, and \(a^*\) and \(b^*\) representing chromaticity. For example, white is \((100, 0, 0)\).
\end{itemize}

In addition, we also tested a combined representation, \textbf{Word(Hex)}. It combines a human-readable color word with its corresponding hex code (\eg, ``white (\#ffffff)''). It provides both the semantic information of color words and the precision of hex codes. Given the estimated representation, we can either take the hex codes (\textit{Word(Hex)-H}) or the words (\textit{Word(Hex)-W}) as the output.

\vspace{-1em}
\subsubsection{Color Conversion} 
Converting between color representations is straightforward when a known conversion formula exists (\eg, \textit{RGB} $\leftrightarrow$ \textit{CIELAB)}. However, converting to or from the \textit{Word} representation is less direct. We addressed this challenge by using the xkcd-color dictionary~\cite{xkcdcolor2024} to map \textit{Hexcode} to \textit{word}. For hex codes without an exact match, we interpolated by identifying the closest RGB value with an associated color word. Similarly, while some color words directly correspond to specific hex codes, most do not appear in the dictionary. To resolve this, we encoded both the missing color words and those in the xkcd-color dictionary using a sentence transformer~\cite{sentence_transformers_all_mpnet_base_v2}, selected the five closest colors, and interpolated their hex values based on the reciprocal of their distances.

\begin{figure}[t]
    \includegraphics[width=1.0\textwidth]{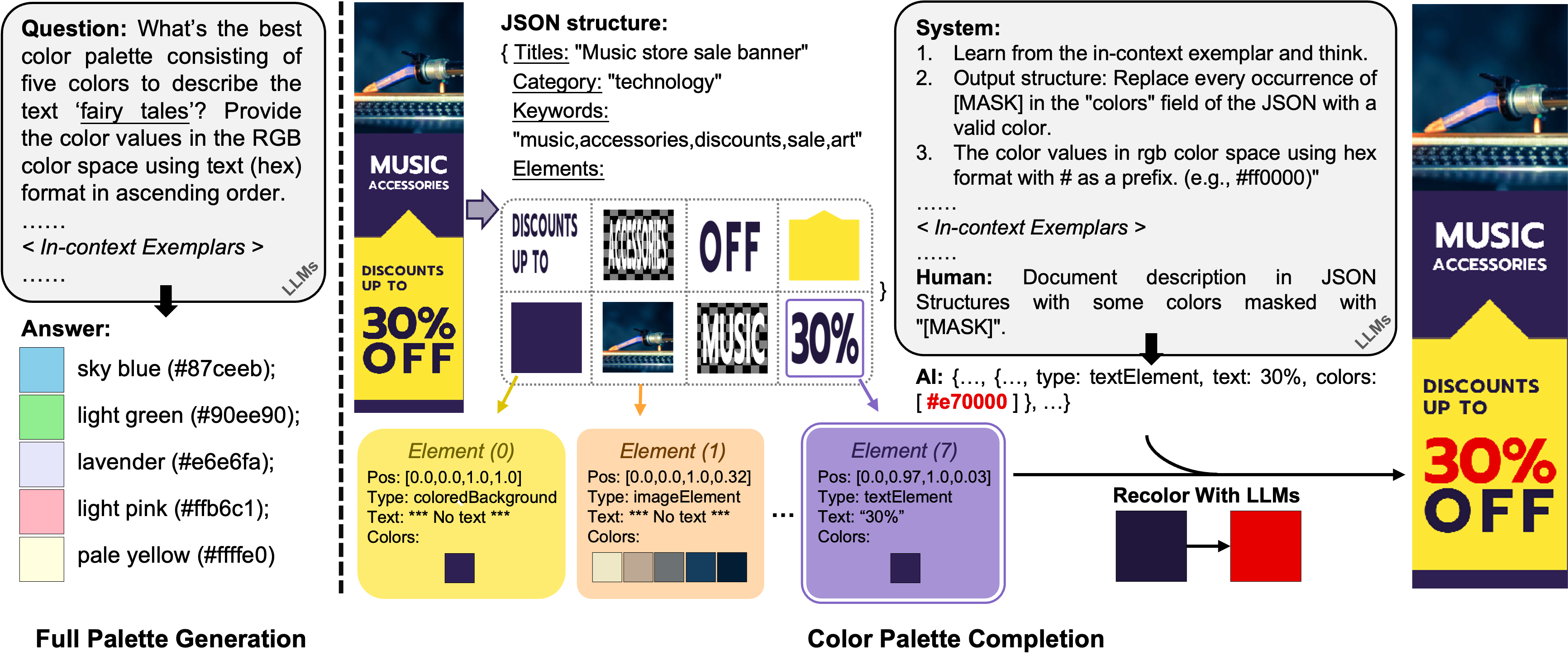}
    \caption{Demonstration of two color recommendation tasks: full palette generation and color palette completion. For the full palette generation, the large language model (LLM) generates a five-color palette from a textual description, with each color shown as ``word(hex)''. In the color palette completion, specified colors are replaced based on surrounding elements and textual prompts, where each element is represented using a JSON structure.}
    \label{fig:pipeline_visualization}
\end{figure}

\subsection{Document Description Structure}
\label{sec:method_structure}
A vector graphic document contains two main types of information: its comprehensive structure and the attributes of its individual elements. The structure includes components such as the theme, keywords, title, and other metadata, while the element attributes cover the various modalities to be rendered, for example, colored backgrounds, scalable vector graphics (SVG) images, raster images (typically in JPEG or BMP format), and text. Additionally, each element includes details required for proper rendering, such as position, size, and opacity. Consequently, organizing this rich information into a format that large language models (LLMs) can understand is crucial for color recommendation in our \texttt{ColorGPT} pipeline.

In methods using Masked Language Models (MLMs)~\cite{qiu2023color,qiu2023multimodal}, inputs are typically short sequences representing color palettes (e.g., ``[red, \_, blue, yellow]'', the blank space ``\_'' indicates a missing color that the model must predict). However, such sequences can be difficult for LLMs to interpret fully, as the meaning of each color is not explicitly defined, often causing the models to copy unmasked colors or rely on familiar examples. To address this limitation, we employed a structured JSON format that explicitly defines the layout and properties of each document element, including title, category, keywords, layout, type, text, and color palette, with the layout specifying the canvas size in unit space so that only overlapping parts are rendered, as illustrated in Fig.\ref{fig:pipeline_visualization}.

\subsection{Prompt Construction}
The prompt design of \texttt{ColorGPT} pipeline can be categorized into three main parts: task-specific profile, output format guidance, and in-context exemplars.

\vspace{-1em}
\subsubsection{Task-specific Profile}
Usually, detailed task-specific profiles can improve the performance of LLMs in specified tasks~\cite{wei2021finetuned,sanh2021multitask,ouyang2022training}. However, in our task, we found LLMs struggled to obtain additional performance increases with extra guidance. On the contrary, simple prompts always have better performance.

\vspace{-1em}
\subsubsection{Output Format Guidance}
In our experiments, we found that omitting the output format description from the prompt often resulted in an unusable model output structure. To address this, we added explicit guidance to ensure that the JSON output consistently adheres to the desired style.

\vspace{-1em}
\subsubsection{In-Context Exemplars}
\label{sec:method_exemplars}
Leveraging in-context learning~\cite{dong2022survey,min2022rethinking}, we guided LLMs using exemplars. For each document, we derived a query from its text and encoded it into word embeddings with a Sentence Transformer~\cite{sentence_transformers_all_mpnet_base_v2}. When presented with a new case, we employed FAISS~\cite{douze2024faiss} to retrieve the most relevant JSON-formatted exemplar from our existing cases, using the top-ranked exemplar for both tasks.

\section{Experiments for Color Palette Completion}
\label{sec:completion}
In this section, we presented the application of \texttt{ColorGPT} to the color palette completion task, which involves suggesting appropriate colors to replace or fill in missing ones in vector graphic documents. The section is organized into three parts. First, we introduced the implementation details of our experiments.
Second, we validated the effectiveness of \texttt{ColorGPT} through a series of experiments. Third, we examined how different component variations impacted the results.

\subsection{Implementation Details}
\label{sec:details_completion}

\subsubsection{Datasets}
For this task, we used the Crello-v2 dataset~\cite{yamaguchi2021canvasvae}, which provides comprehensive document structures and element attributes, including the type of element, position, size, opacity, text contents, and raster images. From the Crello-v2 dataset, Qiu~\etal~\cite{qiu2023multimodal} compiled the MultiPalette-And-Text dataset. We used the same data split to form our JSON structure data. This dataset comprises 14,020, 1,704, and 1,712 valid data points for training, validation, and testing, respectively. 

\vspace{-1em}
\subsubsection{Preprocess}
We categorized graphic document elements into four types: texts, colored backgrounds, SVG images, and raster images. To extract color palettes for each element, we used the method proposed by Chang~\etal~\cite{chang2015palette}, ensuring that each palette contains up to five colors. To maintain diversity, we enforced a minimum Euclidean distance of 10 in the CIELAB color space between any two colors within a palette, preventing overly similar colors from being included. The extracted palettes were then converted to other color representations.

To simulate the need for color completion, we randomly mask one to three colors in the extracted palettes by replacing them with a special token ``[MASK]''. This masking prompts the LLM to infer the missing colors by drawing on contextual cues from the surrounding colors and the overall palette design.

\vspace{-1em}
\subsubsection{Baselines}
We compared the performance of ColorGPT with two related works. One recommends colors based on surrounding colors~\cite{qiu2023color}, while the other incorporates text embeddings into the color recommendation model~\cite{qiu2023multimodal}. We adopted \textit{Hexcode} representation and used GPT-4o (2024-08-06)~\cite{achiam2023gpt} in this experiment.

\vspace{-1em}
\subsubsection{Recoloring}
\label{subsubsec:recoloring}


After obtaining the suggested color, we applied recoloring using the method proposed by Chang~\etal~\cite{chang2015palette}. Baseline approaches assigned colors based on broad groups such as images, graphics, or texts, so when one color was provided, it often affected multiple elements within that group. In contrast, our method assigned a unique color palette to each element, ensuring that only the targeted element is recolored and preserving detailed color information.

\vspace{-1em}
\subsubsection{Evaluation Metrics}
To assess the color palette completion capabilities of our approach, we employed two evaluation metrics for \texttt{ColorGPT} and the baselines.

First, we used an \textit{Accuracy} metric that compares the predicted colors against the ground truth. Following the method of Qiu~\etal~\cite{qiu2023multimodal}, all colors were quantized into 16×16×16 bins. For cases where only one color is suggested, we applied the same quantization process to the output of \texttt{ColorGPT}. A prediction is considered accurate if the suggested color falls within the same bin as the ground truth, and overall accuracy was calculated as the percentage of correct predictions. In scenarios that require the suggestion of multiple colors (e.g., two or three), a case was considered correct only if every suggested color matches the corresponding ground-truth color bin.

Second, we introduced a \textit{Distribution}~\cite{bahng2018coloring} metric to ensure that the model's color suggestions are not biased toward certain colors. We observed in our experiments that black and white frequently appear in design data, so it is crucial to verify that improvements in accuracy are not disproportionately influenced by these two colors. A higher distribution value in the predicted colors indicates a broader, more balanced range of suggestions, which in turn demonstrates better diversity in the model's predictions.

\subsection{Results}
\label{sec:results_completion}
\subsubsection{Color Completion Quality}
\begin{figure}[t]
    \centering
    \includegraphics[width=\textwidth]{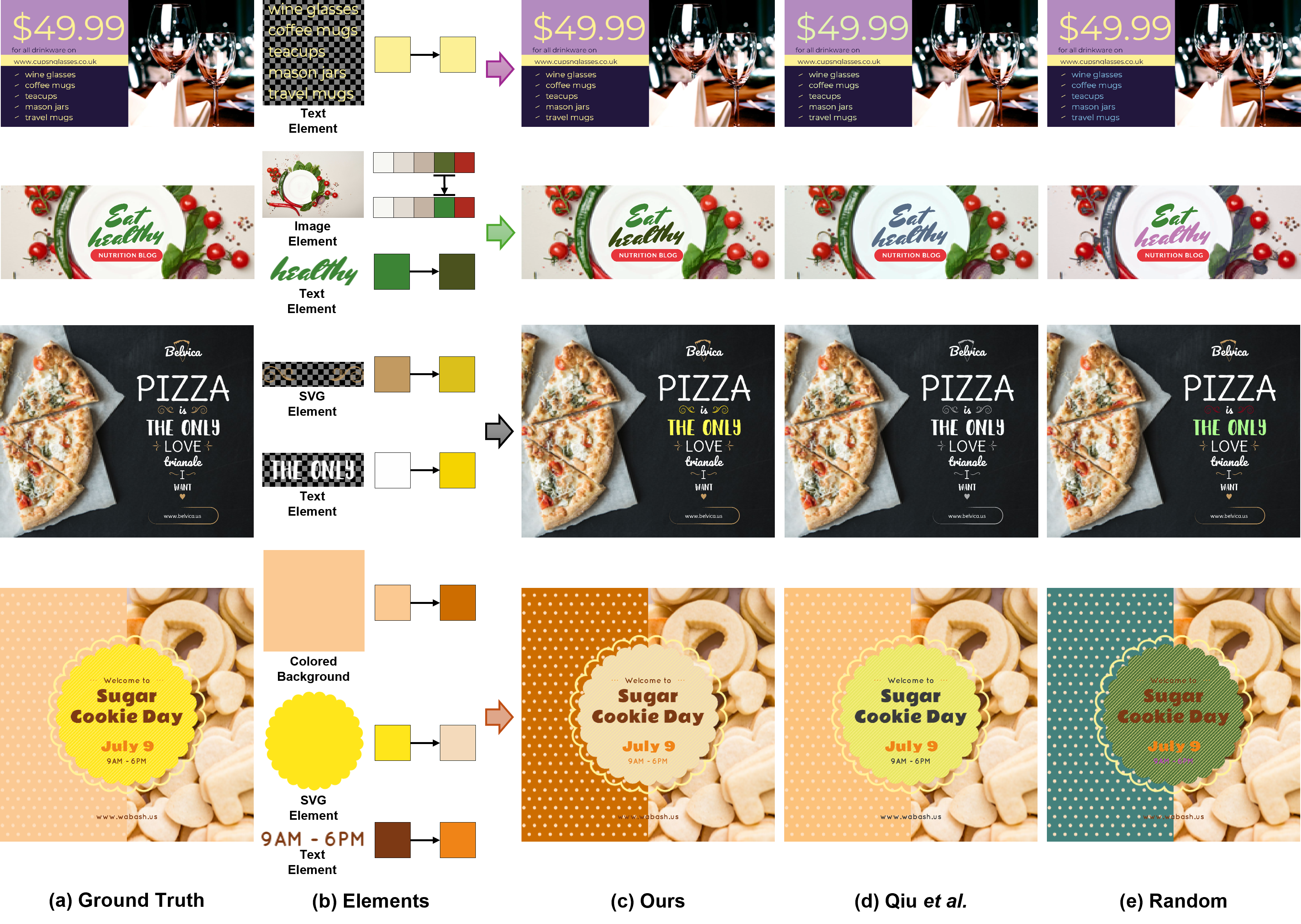} 
    \vspace{-2em}
    \caption{Color recommendation results with our method, Qiu~\etal~\cite{qiu2023multimodal}, and random color completion. All these examples were from the test set. The elements requiring color suggestion were selected randomly.}
    \label{fig:comparison_completion}
\end{figure}

Table~\ref{tab:accuracy_completion} compared \texttt{ColorGPT} with baseline methods and the ground truth, while Fig.~\ref{fig:comparison_completion} visually illustrated several cases.

First, as shown in Table~\ref{tab:accuracy_completion}, \texttt{ColorGPT} achieved superior accuracy over all masked color numbers compared to the baseline methods. This performance was especially impressive given that the color suggestions were generated using the model's general knowledge, a simple task-specific profile, and one in-context exemplar.
In addition, we observed that as the number of masked colors increased, the performance gap between our method and the baseline became larger. This effect was due to the differing recoloring procedures employed by the two methods (see Sec.~\ref{sec:details_completion} Recoloring). For the baseline method, as the number of masked colors increased, the number of affected elements increased significantly, resulting in a more pronounced drop in performance.

\begin{table}[t]
  \centering
  \setlength{\tabcolsep}{5pt} 
  \small
  \caption{Comparing accuracy and distribution with baseline methods and ground truth. Methods marked with {\textdagger} are supervised models.}
  \label{tab:accuracy_completion}
  \begin{tabular}{lcccccc}
    \toprule
    \multirow{2}{*}{\textbf{Method}} & \multicolumn{3}{c}{\textbf{Accuracy(\%)$\uparrow$}} & \multicolumn{3}{c}{\textbf{Distribution$\uparrow$}} \\
    & 1-color & 2-color & 3-color & 1-color & 2-color & 3-color \\
    \midrule    
    Qiu~\etal~\cite{qiu2023color}\textsuperscript{\textdagger} & 36.72 & 16.04 & 6.45 & 4.93 & 4.31 & 2.94 \\
    Qiu~\etal~\cite{qiu2023multimodal}\textsuperscript{\textdagger}  & 47.13 & 26.22 & 15.67 & 6.04 & 6.64 & 6.30 \\
    \midrule
    ColorGPT & \textbf{52.60} & \textbf{31.64} & \textbf{22.61} & 4.34 & 4.53 & 4.54 \\
    \midrule
    Ground Truth & - & - & - & 4.49 & 4.56 & 4.58 \\
    \bottomrule
  \end{tabular}  
\end{table}

In Fig.~\ref{fig:comparison_completion}, we presented color completion results for one-, two-, and three-color cases applied to various elements. For comparison, we included the original design (Ground Truth), outputs from our method, the baseline method by Qiu~\etal~\cite{qiu2023multimodal}, and a random color completion, which were all taken from the test set. Our results showed that our method provided color suggestions that aligned with the overall theme of the vector graphic documents and enhanced the harmony among surrounding elements. Compared to random color assignments, our approach produced more visually coherent palettes, and its performance was comparable to that of Qiu~\etal~\cite{qiu2023multimodal}. Moreover, for the second case, our method offered the unique advantage of allowing individual color adjustments for each element, unlike the statistical palette approach used by Qiu~\etal~\cite{qiu2023multimodal}.

Secondly, our evaluation of the distribution metric showed that \texttt{ColorGPT} exhibits a variance similar to that of the ground truth. This indicated that \texttt{ColorGPT} shares a color preference closely aligned with the ground truth, which we examined in detail in the following section.

Overall, these results indicated that \textit{pretrained LLMs have considerable potential as effective designers for the color palette completion task}. They demonstrated adaptability for graphic documents, maintaining color harmony even in complex designs. However, further analysis of \texttt{ColorGPT}'s reference was needed.

\vspace{-1em}
\subsubsection{Suggested Color Preference}
\begin{figure}[t]
    \centering
    \begin{tabular}{cc}
    \includegraphics[width=0.3\textwidth]{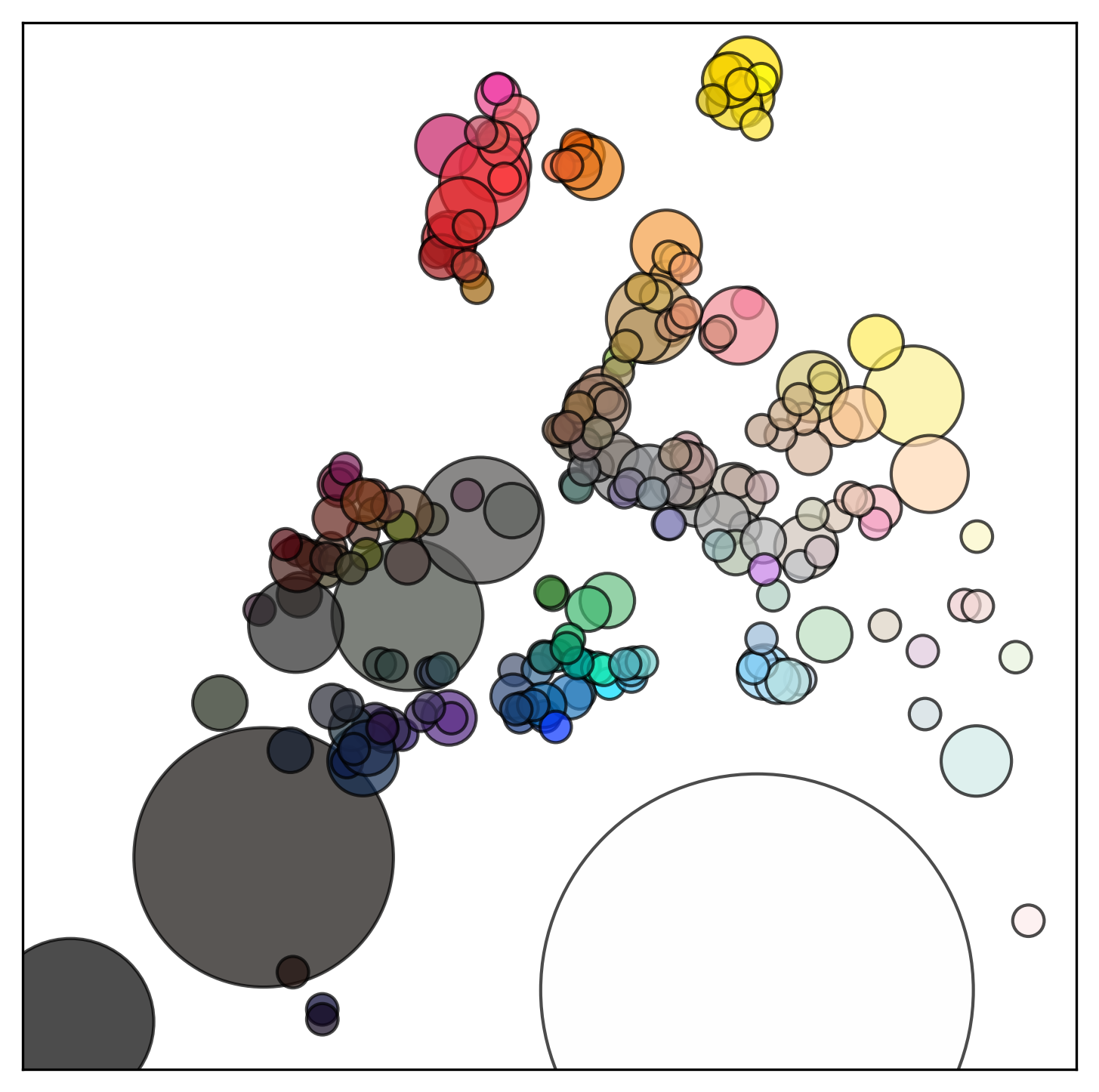} &
    \includegraphics[width=0.3\textwidth]{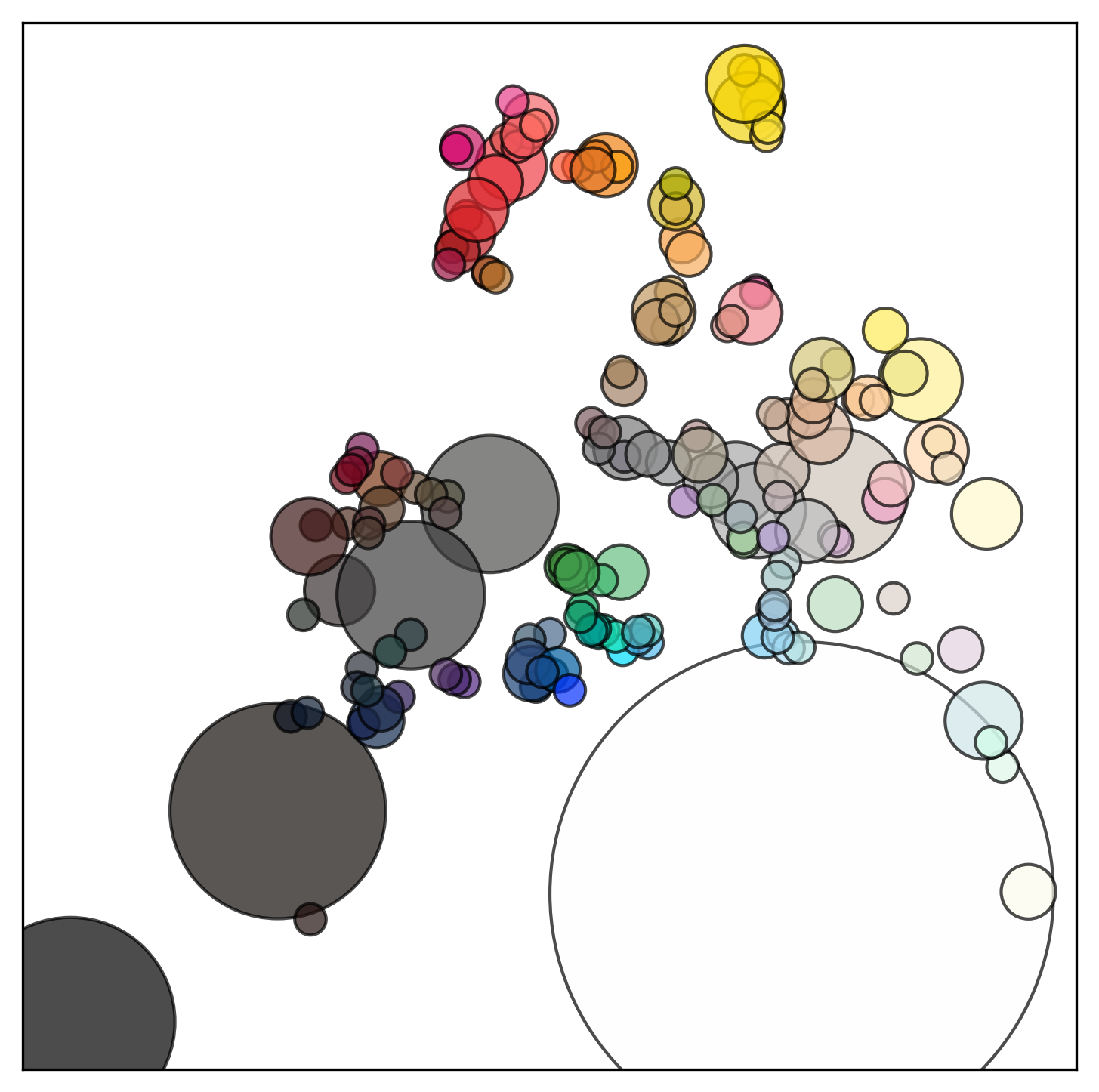} \\
    (a) Ground Truth  &  (b) \texttt{ColorGPT}
    \end{tabular}
    \caption{Color distribution that the data point is assigned with its own color and the point size reflects its frequency.}
    \label{fig:tsne_completion}
\end{figure}

In the previous section, we observed that LLMs have similar color preferences as the ground truth. To investigate further, we visualized both the predicted colors and the ground truth using t-SNE (see Fig.~\ref{fig:tsne_completion}). The comparison validated that the overall distribution of predicted colors largely mirrors that of the ground truth. However, the t-SNE plot reveals small differences in the regions corresponding to white and black, suggesting that \texttt{ColorGPT} tended to favor white slightly over other colors, which indicates a potential bias in its color suggestions.

\begin{table}
  \centering
  \setlength{\tabcolsep}{5pt} 
  \small
  \caption{Accuracy of color completion task across different element types. We compared only 1-color mask in this table.}
  \label{tab:element_accuracy_completion}
  \begin{tabular}{lccc}
    \toprule
    \multirow{2}{*}{\textbf{Method}} & \multirow{2}{*}{\textbf{Ratio(\%)}} & \multicolumn{2}{c}{\textbf{Accuracy(\%)$\uparrow$}}\\ 
    & & \texttt{ColorGPT} & Qiu~\etal~\cite{qiu2023multimodal} \\
    \midrule
    Text & 29.0 & 71.74 & 66.95 \\
    SVG & 38.7 & 54.91 & 52.93 \\
    Raster Image & 32.3 & 28.16 & 37.79 \\
    \midrule
    ColorGPT & - & 52.60 & 47.13 \\
    \bottomrule
  \end{tabular}
\end{table}

\vspace{-1em}
\subsubsection{Element-wise Analysis}
In this section, we evaluated the accuracy of different element types to better understand their contributions to overall performance (see Table~\ref{tab:element_accuracy_completion}). Our analysis revealed that raster image elements achieved significantly lower accuracy compared to other types. We hypothesized that this discrepancy arises from the limited five-color palette used in raster images, which restricted the models’ ability to capture and interpret their inherent complexity. In contrast, simpler elements such as text and SVG images achieved higher accuracy due to their straightforward structures, which were easier for the models to process. These findings suggested that for complex elements like raster images, textual descriptions alone may be insufficient for effective understanding. Future research should consider employing multi-modal LLMs to further improve the performance of color palette recoloring.

\begin{table}[t]
  \centering
  \setlength{\tabcolsep}{5pt} 
  \small
  \caption{Component variation discussion of \texttt{ColorGPT} on four aspects: color representation, task-specific profile length, w./wo. JSON structure and exemplar types. {\textdagger} is the default setting of \texttt{ColorGPT}.}
  \label{tab:completion_ablation}
  \begin{tabular}{lccc}
    \toprule
     \multirow{2}{*}{\textbf{Component Variation}} & \multicolumn{3}{c}{\textbf{Accuracy(\%)$\uparrow$}} \\
    & 1-color & 2-colors & 3-colors \\
    \midrule
    \multicolumn{4}{l}{\textit{Color Representations}} \\
    ~~~~Hexcode\textsuperscript{\textdagger} & \textbf{52.60} & \textbf{31.64} & \textbf{22.61} \\
    ~~~~Word & 17.06 & 6.23 & 2.71 \\
    ~~~~RGB & 42.86 & 26.02 & 19.18 \\
    ~~~~CIELAB & 38.78 & 25.12 & 15.23 \\
    ~~~~Word(Hex)-H & 42.86 & 28.31 & 17.72 \\
    ~~~~Word(Hex)-W & 18.47 & 7.67 & 2.70 \\
    \midrule
    \multicolumn{4}{l}{\textit{Task-specific Profile Length}} \\
    ~~~~short\textsuperscript{\textdagger} & \textbf{52.60} & \textbf{31.64} & \textbf{22.61} \\
    ~~~~long & 23.40 & 9.87 & 5.43  \\
    \midrule
    \multicolumn{4}{l}{\textit{w./wo. JSON Structure}} \\
    ~~~~w. JSON\textsuperscript{\textdagger} & \textbf{52.60} & \textbf{31.64} & \textbf{22.61} \\
    ~~~~wo. JSON & 30.43 & 11.57 & 7.01 \\
    \midrule
    \multicolumn{4}{l}{\textit{Exemplar Types}} \\
    ~~~~ Random & 31.60 & 11.15 & 4.76 \\
    ~~~~ Similarity\textsuperscript{\textdagger} & \textbf{52.60} & \textbf{31.64} & \textbf{22.61} \\
    \bottomrule
  \end{tabular}
\end{table}

\subsection{Discussion}
\label{sec:discuss_completion}

\subsubsection{Different Component Variation}
First, As shown in Table~\ref{tab:completion_ablation}, \textit{Hex} achieved the best accuracy, with \textit{RGB} and \textit{CIELAB} ranking second and fourth, respectively, indicating that Hex is the optimal color representation for this task. In contrast, the \textit{Word} and \textit{Word(Hex)-W} methods performed the worst, likely because the color conversion process (Sec.~\ref{sec:method_color}) impaired their ability to suggest color palettes accurately.
Additionally, we also tested a combined representation, Word(hex), to improve results, the additional color word information ultimately reduced the overall accuracy of \textit{Word(Hex)-W}.

Second, we found that including detailed instructions in the task-specific profile had the opposite effect --- reducing the diversity of suggestions. We believe this is because our JSON structure already provided sufficient information for color palette completion. In effect, the extra detailed instructions might constrain the LLM’s design flexibility, which may undermine performance by limiting its ability to fully leverage LLMs' knowledge.

Third, we compared performance with and without a JSON structure. In the version without a JSON structure, we designed an approach guided by MLMs (Sec.~\ref{sec:method_structure}) that allowed LLMs to output only the color, rather than generating a complete JSON with replaced masked colors. The results showed that using a JSON structure significantly improved performance.

Fourth, we compared two in-context exemplar sampling methods, random sampling, and similarity-based selection, and found that similarity-based exemplars, which better reflect the task context, improved accuracy.

\begin{figure}[t]
    \centering
    \includegraphics[width=\textwidth]{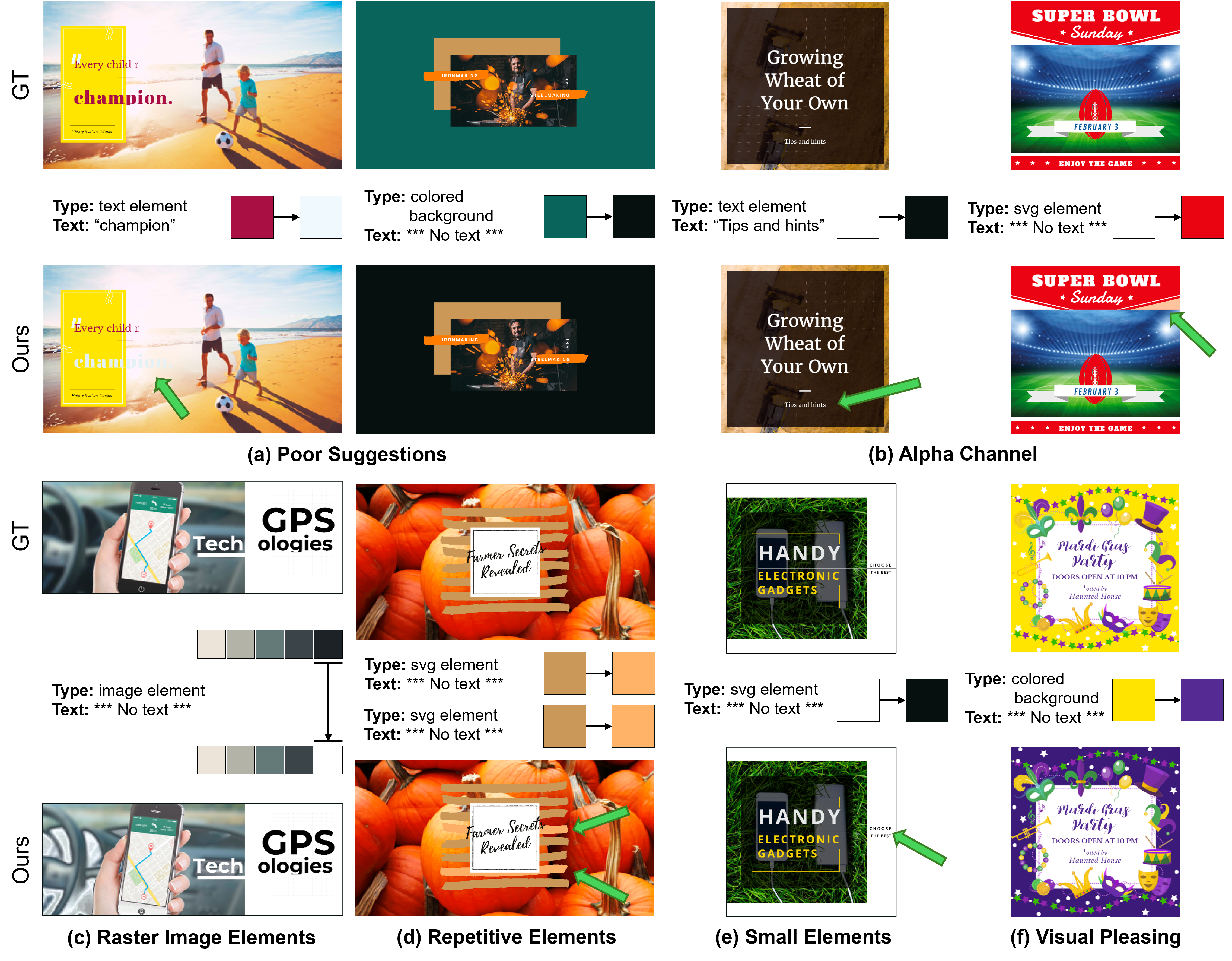}
    \vspace{-2em}
    \caption{Analysis of failed cases. We summarized the failure cases into six main types.}
    \label{fig:completion_failure}
\end{figure}

\vspace{-1em}
\subsubsection{Failure Case Analysis}
Fig.~\ref{fig:completion_failure} illustrated six types of failures observed in the color palette completion task.
For the poor suggestion case (Fig.~\ref{fig:completion_failure}(a)), \texttt{ColorGPT} sometimes failed to ensure sufficient contrast with the background. This issue may be rooted in the limited information available for raster image elements.
In the alpha channel issue (Fig.~\ref{fig:completion_failure}(b)), some elements from the Crello-V2 dataset were in RGBA format. Hence, the RGB values did not fully capture the final rendered colors because the alpha channel influenced the appearance. As shown in the figure, the model recommended changing a text element to black; however, it appeared white in the final rendering due to its transparency. An improvement might involve extracting the palette from the fully rendered output instead of relying solely on the raw RGBA data.
Regarding raster image elements (Fig.\ref{fig:completion_failure}(c)), as discussed in Sec.\ref{sec:results_completion}, completing colors for raster images remained challenging, especially when the model was provided with only up to five colors to make color suggestions.
For repetitive elements (Fig.~\ref{fig:completion_failure}(d)), the model often proposed colors that are close, but not identical, to the existing ones. This slight variation can lead to visual inconsistency and discomfort, likely because the model struggles to fully recognize and replicate repetitive structures.
In the case of small elements (Fig.~\ref{fig:completion_failure}(e)), the model appeared to have difficulty discerning their contextual relationship with surrounding elements, which resulted in recoloring failures.
Interestingly, we also observed instances (Fig.~\ref{fig:completion_failure}(f)) where the generated colors deviated from the ground truth but produced visually pleasing results. This suggested that the model might offer creative color suggestions beyond strict adherence to the original palette.

\section{Experiments for Full Palette Generation}
\label{sec:exp_generation}
In this section, we explored \texttt{ColorGPT}'s application in full palette generation, a task in which a complete color palette is generated based solely on a given text. This experiment highlighted the model's ability to generate effective palettes using only contextual information.

\subsection{Implementation Details}
\label{sec:detail_generation}

\subsubsection{Datasets}
For the experiments, we adopted the manually curated Palette-And-Text (PAT) dataset~\cite{bahng2018coloring}. This dataset contains 10,183 pairs, each comprising a five-color palette and its corresponding textual description. The descriptions are closely linked to the colors: some words directly refer to color names (e.g., `pink', `grey'), while others evoke specific color themes (e.g., `summer', `autumn'). We randomly divided the dataset into 8,147 training pairs, 1,018 validation pairs, and 1,018 testing pairs, following the experimental setup of previous work~\cite{qiu2023multimodal}. The training pairs were subsequently used to extract exemplars during the inference process.

\vspace{-1em}
\subsubsection{Evaluation Metrics}
We employed two complementary metrics that capture distinct aspects of palette quality. First, the \textit{Similarity} metric measures how closely the predicted palette matches the ground truth. To compute this, we used Dynamic Closest Color Warping (DCCW)~\cite{kim2021dynamic}, which calculates the minimal distance between corresponding colors in the generated and ground truth palettes, thereby quantifying their visual closeness. Second, the \textit{Diversity} metric assesses the average pairwise distance among the five colors in each palette, following the approach used in prior works~\cite{bahng2018coloring,qiu2023multimodal}. This metric revealed the variation within the generated palette, ensuring that it does not comprise overly similar colors, which would diminish its usefulness.

\vspace{-1em}
\subsubsection{Baselines}
We evaluated \texttt{ColorGPT} by comparing it with two methods trained on task-specific datasets. The first method, Text-to-Palette Generation Networks (TPN)~\cite{bahng2018coloring}, employed a conditional GAN as detailed in the original work. We also compared our model with the method proposed by Qiu \etal~\cite{qiu2023multimodal}. For this task, we utilized GPT-4o-mini (2024-07-18)~\cite{achiam2023gpt} with a \textit{Word(Hex)} representation, extracting only the hex codes to compute the evaluation metrics.

\subsection{Results}
\label{sec:result_generation}

\subsubsection{Generation Quality}
\begin{figure}[t]
    \centering
    \hspace*{-10pt}
    \includegraphics[width=0.92\textwidth]{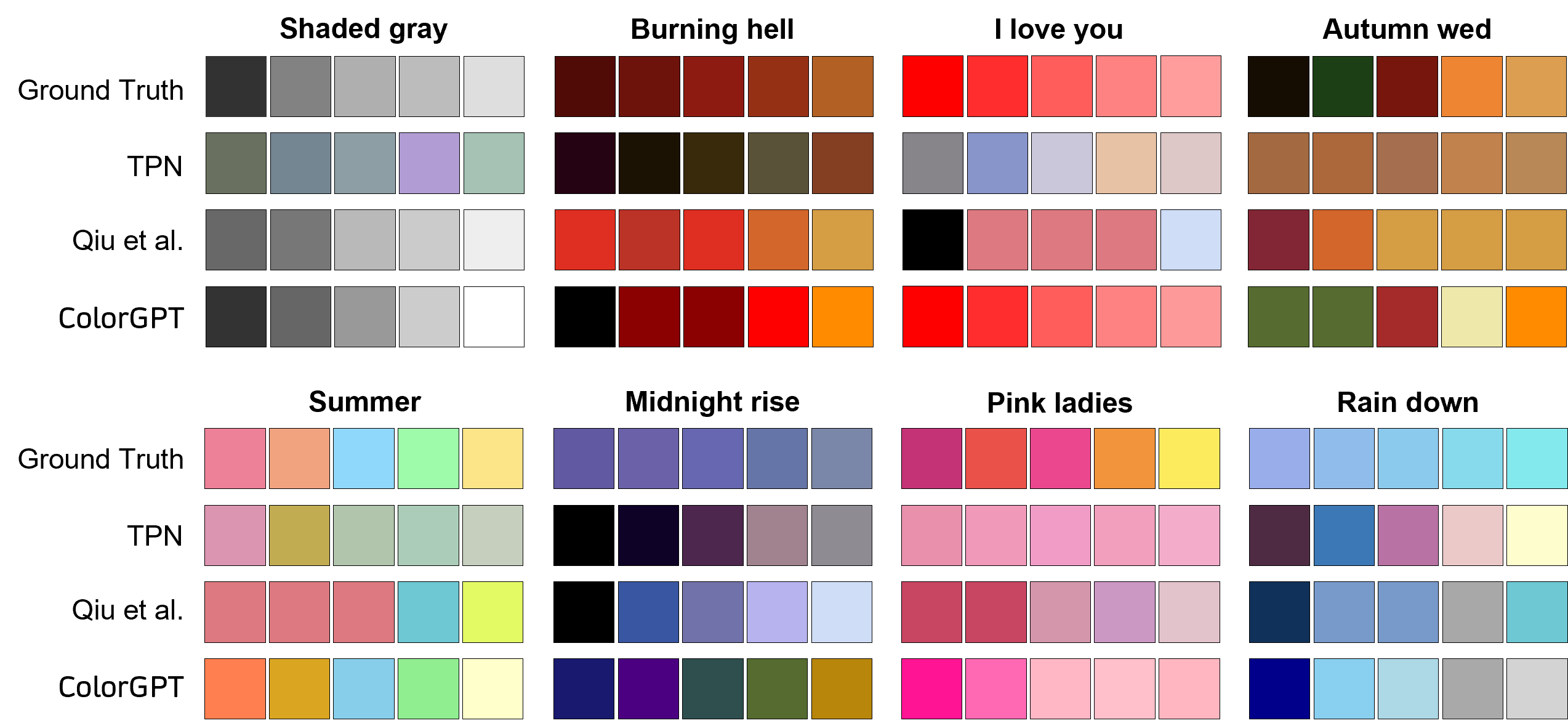}
    \caption{\looseness=-1
    Qualitative analysis on textual context. We compare the generated palette results of ColorGPT, the related work TPN~\cite{bahng2018coloring} and Qiu~\etal~\cite{qiu2023multimodal} with the ground truth.}
    \label{fig:palette_generation}
\end{figure}

Table~\ref{tab:quality_generation} compared the average and standard deviation of two metrics, similarity and diversity, with state-of-the-art methods. And Fig.~\ref{fig:palette_generation} presents visual comparisons of selected cases.
In Table~\ref{tab:quality_generation}, we found that \texttt{ColorGPT} outperformed other methods in similarity. This result demonstrated that LLMs can effectively recommend color palettes based solely on contextual descriptions. Regarding diversity, \texttt{ColorGPT} generally achieved scores very close to the ground truth, which indicated that LLMs are able to generate palettes with a broader range of colors rather than only recommending similar hues.
This trend was further illustrated in Fig.~\ref{fig:palette_generation}. The recommendations from \texttt{ColorGPT} were more aligned with the ground truth compared to baseline methods. In addition, our method produced palettes with greater variety and effectiveness, while baseline approaches tended to suggest colors within a narrow tonal range.

Together, these findings validated that \textit{pretrained LLMs can serve as superior designers for full palette generation tasks.} However, further analysis was needed.

\begin{table}[t]
  \centering
  \small
  \setlength{\tabcolsep}{5pt} 
  \caption{Quantitative analysis results of palette similarity to ground truth and color diversity. In Qiu's work~\cite{qiu2023multimodal}, a post-processing (PP) step was adopted to eliminate duplicated colors in a palette. We compared the without-pp version for fairness.}
  \label{tab:quality_generation}
  \begin{tabular}{lcccc}
    \toprule
    \multirow{2}{*}{\textbf{Method}} & \multicolumn{2}{c}{\textbf{Similarity$\downarrow$}} & \multicolumn{2}{c}{\textbf{Diversity$\uparrow$}} \\
    & Average & Std & Average & Std \\
    \midrule
    TPN~\cite{bahng2018coloring} & 29.26 & 13.35 & 22.21 & 10.78 \\
    Qiu~\etal(w/o pp)~\cite{qiu2023multimodal} & 28.14 & 12.91 & 29.92 & 10.27 \\
    \texttt{ColorGPT} & \textbf{26.09} & 11.42 & 35.80 & 11.76 \\
    \midrule
    Ground Truth & - & - & 26.17 & 13.84 \\
    \bottomrule
  \end{tabular}
\end{table}

\vspace{-1em}
\subsubsection{Color Representations}
We evaluated various color representations (Table~\ref{tab:variation_generation}). Among them, the \textit{Word(Hex)-H} format outperformed the others because it combines the semantic richness of \textit{Word} with the precision of \textit{Hexcode}, thereby enhancing the LLMs’ ability to manage color tasks. Although not the top performer, \textit{Word(Hex)-W} also demonstrated that integrating two color representations can improve performance. The advantage of \textit{Word(Hex)-H} may stem from the interpolation process during the transformation of color words into RGB values, which is an essential consideration since many color words lack exact RGB equivalents (see Sec.~\ref{sec:method_color}). Expanding the color dictionary could potentially enhance the performance of both the \textit{Word} and \textit{Word(Hex)-W} representations.
In addition, when using a single representation, the \textit{Word} yielded the best results, while \textit{CIELAB} performed the worst. We attributed this outcome to the frequency of these representations in the training corpus: \textit{Word} representations are more common in everyday language, whereas \textit{CIELAB} values are less frequently encountered and are typically reserved for specialized professional contexts. 

Earlier observations indicated that for the color palette completion task, \textit{Hexcode} offered the precise representation needed for accurate evaluation. In contrast, the results of the full palette generation task showed that the best representation was \textit{Word(Hex)-H}. For completion, where precise prediction from surrounding colors is critical, \textit{Hexcode} is superior due to its clear and unambiguous format. Meanwhile, full palette generation involves linking descriptive language to color names, which is a subjective process that may not need exact color accuracy. Thus, we conjecture that adding color names as an auxiliary representation helped LLMs better grasp the task, boosting the performance.

\begin{table}[t]
  \centering
  \setlength{\tabcolsep}{5pt} 
  \small
  \caption{
  Comparison of \texttt{ColorGPT} component variations for full palette generation task.}
  \label{tab:variation_generation}
  \begin{tabular}{lcccc}
    \toprule
    \multirow{2}{*}{\textbf{Component Variation}} &  \multicolumn{2}{c}{\textbf{Similarity$\downarrow$}} & \multicolumn{2}{c}{\textbf{Diversity$\uparrow$}} \\
     & Mean & Std & Mean & Std \\
    \midrule
    \textit{Model Sizes} & & & & \\
    ~~~~Llama3-8B & 33.34 & 14.63 & 34.51 & 13.62 \\
    ~~~~GPT3.5-turbo & 27.93 & 12.59 & 34.96 & 13.28 \\
    ~~~~GPT4o\textsuperscript{\textdagger} & \textbf{26.09} & 11.42 & 35.80 & 11.76  \\ 
    \midrule
    \textit{Color Representations} & & & & \\
    ~~~~Word & 32.61 & 14.92 & 37.29 & 12.92 \\
    ~~~~Hexcode & 34.83 & 19.27 & 27.44 & 14.46 \\
    ~~~~RGB & 35.54 & 17.40 & 37.18 & 17.35 \\
    ~~~~CIELAB & 39.42 & 20.22 & 17.30 & 13.58 \\
    ~~~~Word(Hex)-W & 28.08 & 12.14 & 37.73 & 13.21 \\
    ~~~~Word(Hex)-H\textsuperscript{\textdagger} & \textbf{26.09} & 11.42 & 35.80 & 11.76  \\ 
    \midrule
    \textit{Exemplar Types} & & & & \\
    ~~~~Random & 28.50 & 14.36 & 28.71 & 13.51 \\ 
    ~~~~Similarity\textsuperscript{\textdagger} & \textbf{26.09} & 11.42 & 35.80 & 11.76  \\ 
    \bottomrule
  \end{tabular}
\end{table}

\vspace{-1em}
\subsubsection{In-context Exemplars.}
We compared two in-context exemplar sampling methods, random sampling versus similarity-based selection for the full palete generation task. Similar to the previous task, we found that similarity-based exemplars, which better reflect the task context, resulted in improved accuracy.

\section{Conclusion}
\label{sec:conclude}
We introduced \texttt{ColorGPT}, a training-free method that leverages LLMs for color design in vector graphics. Focusing on palette completion and generation, we showed LLMs’ commonsense reasoning is effective. We analyzed key performance factors and proposed an enhanced document representation to boost LLM comprehension. Our results indicate that \textit{hexcode} serve as a strong baseline for color representation, and adding descriptive words further improves accuracy. As LLMs continue to advance, our research extended their application to color recommendation, offering the potential to facilitate designers’ workflows.

%
%
%
\bibliographystyle{splncs04}
\bibliography{mybibliography}

\end{document}